\documentclass[sigconf]{acmart}

\usepackage{stfloats}
\usepackage{algorithm}
\usepackage{algorithmic}
\usepackage{mathrsfs}
\usepackage{makecell}
\usepackage{array}
\usepackage{enumitem}
\usepackage{siunitx}
\usepackage{subfig}
\newcommand{\subfigure}{\subfloat}
\newcommand{\yang}[1]{}
\newcommand{\jun}[1]{}
\newcommand{\yihang}[1]{}
\newcommand{\zhongwei}[1]{}

\acmYear{2026}
\acmConference[DAI '26]
  {The Eighth International Conference on Distributed Artificial Intelligence}
  {November 29--December 2, 2026}
  {Hong Kong, China}
\acmBooktitle{The Eighth International Conference on Distributed Artificial
  Intelligence (DAI '26), November 29--December 2, 2026, Hong Kong, China}

\setcopyright{none}
\copyrightyear{}
\acmDOI{}
\acmISBN{}

\begin{document}

\title{Embodied Science Agents: Making Embodied Experimental Design Learnable}

\author{Yihang Chen}
\authornote{Equal contribution.}
\affiliation{%
  \institution{University College London}
  \city{London}
  \country{United Kingdom}}
\affiliation{%
  \institution{AI Lab, The Yangtze River Delta}
  \country{China}}

\author{Zhongwei Yu}
\authornotemark[1]
\affiliation{%
  \institution{The Hong Kong University of Science and Technology (Guangzhou)}
  \city{Guangzhou}
  \country{China}}
\affiliation{%
  \institution{AI Lab, The Yangtze River Delta}
  \country{China}}

\author{Yang Li}
\authornotemark[1]
\authornote{Corresponding authors.}
\affiliation{%
  \institution{Shanghai Jiao Tong University}
  \city{Shanghai}
  \country{China}}
\email{yang.li.cs@sjtu.edu.cn}

\author{Xingyu Lu}
\affiliation{%
  \institution{The Hong Kong University of Science and Technology (Guangzhou)}
  \city{Guangzhou}
  \country{China}}
\affiliation{%
  \institution{AI Lab, The Yangtze River Delta}
  \country{China}}


\author{Hongxia Yu}
\affiliation{%
  \institution{Nanjing Agricultural University}
  \city{Nanjing}
  \country{China}}

\author{Jun Wang}
\authornotemark[2]
\affiliation{%
  \institution{University College London}
  \city{London}
  \country{United Kingdom}}
\email{junwang@cs.ucl.ac.uk}

\renewcommand{\shortauthors}{Chen et al.}

\begin{abstract}
Recent advances in large models have accelerated AI for Science toward agentic science, where AI agents autonomously conduct scientific discovery. Realising this vision requires closing the loop between computational reasoning and physical experimentation, positioning self-driving laboratories (SDLs) as critical enablers. However, existing SDLs exhibit a fundamental paradox: foundation model integration has increased linguistic sophistication while severing physical grounding. By treating models as disembodied planners that output symbolic workflows for separate robotic systems, current architectures inherently block the development of embodied experimental intuition.
We argue that overcoming this paradox requires embedding \emph{learnability} along three interdependent axes: cognitive learnability (learning what to investigate), experimental learnability (learning where to search), and embodied learnability (learning how to act). Building on this view, we introduce the Embodied Science Agent (ESA), which unifies digital intelligence with physical embodiment through three synergistic components: a Digital Agent that provides cognitive and experimental learnability via memory-based reasoning and Bayesian optimisation, an Embodied Agent that provides embodied learnability via generalisable instrument control, and an Embodied Unified Interface for seamless coordination.
We validated ESA in autonomous wastewater treatment experiments, navigating over 200 million possible conditions. Through 15 continuous rounds, ESA raised the final solution transmittance from 82\% to 99.7\%, corresponding to a 98.5\% reduction in residual dye concentration and near-complete decolourisation, while its learned visuomotor policies generalised across varying instrument placements, a promising step toward next-generation SDLs for autonomous scientific discovery.

\end{abstract}



\maketitle

\section{Introduction}
Artificial intelligence is transforming scientific discovery, driving AI for Science~(AI4S) as a pivotal frontier~\cite{jumper2021alphafold,yang2023alphafold2,noe2019boltzmann}. Within it, self-driving laboratories~(SDLs) couple computational reasoning with robotic experimentation to actively test hypotheses in the physical world, closing the loop between modelling and empirical validation~\cite{Burger2020MobileRoboticChemist,sdlsurvey}. The recent integration of foundation models has expanded SDLs from narrow, protocol-driven platforms into generalisable scientific agents that reason autonomously across chemistry, materials science, and biology~\cite{pyzerknapp2025foundation,choi2025perspective,guo2025foundation,menon2025scientificFMs,hu2025survey}.

Yet these advances mask a critical gap: today's SDLs think about experiments, but do not experience them.
Human experimental expertise is fundamentally embodied: through repeated sensorimotor interaction, chemists and biologists develop tacit sensitivity to cues such as colour gradients, reaction textures, or fluid resistance during pipetting, which enables techniques to transfer across reagents and instruments and to adapt when conditions deviate from written protocols. Current SDLs, by contrast, maintain a strict division between foundation models for reasoning and robotic systems for execution, precluding embodied expertise and yielding three systemic limitations: insufficient physical grounding (operations treated as symbolic actions devoid of material dynamics), limited adaptivity (no real-time feedback between planning and execution), and absence of sensorimotor learning (no accumulation of experimental intuition). Closing this gap requires embodied scientific agents that unify perception, cognition, and action to learn through physical interaction.

To address these limitations, we argue that next-generation SDLs must embed \emph{learnability} along three interdependent axes that together constitute embodied scientific intelligence: \textbf{cognitive learnability}, learning \emph{what} to investigate by reflecting on past reasoning; \textbf{experimental learnability}, learning \emph{where} to search by sequentially refining experimental designs from evidence; and \textbf{embodied learnability}, learning \emph{how} to act by acquiring robust manipulation skills through interaction. These axes are deeply interdependent: genuinely autonomous discovery emerges from their co-evolution rather than from optimising any one in isolation (formalised in Appendix~\ref{app:learnability}).

Guided by this vision, we introduce the Embodied Science Agent~(ESA), a new embodied self-driving laboratory paradigm that transforms SDLs from automated experimenters into adaptive, self-evolving agents. ESA couples digital intelligence with physical systems through three tightly coupled components: (i)~a \textbf{Digital Agent}, the cognitive core (cerebrum) driving open-ended evolution via adaptive reasoning, Bayesian optimisation (BO), and continual learning from memory; (ii)~an \textbf{Embodied Agent} (cerebellum) that unifies diverse robotic and scientific instruments through multiple embodied control modules for generalised, reliable operation; and (iii)~an \textbf{Embodied Unified Interface (EUI)} providing a standardised, low-latency, safe communication protocol that bridges the two across heterogeneous instruments. Internally, ESA is realised as a distributed multi-agent system in which specialised agents---a Principal Investigator agent for planning, a Bayesian optimiser for experimental design, perception and GUI agents for instrument interaction, an embodied control agent, and a reporter agent---cooperate through the EUI to close the perception--reasoning--action loop. Grounding such distributed agents in real physical experimentation, and coordinating them safely and reliably, is precisely the class of problem central to distributed AI (DAI); ESA offers a concrete testbed where distributed cognition must be tightly coupled with embodied action.

To evaluate this paradigm, we implemented ESA in a functional chemical laboratory that integrates conventional instruments with a dual-arm embodied platform. The system autonomously conducted a complete optimisation cycle for the Electro-Fenton degradation of Safranine T wastewater---planning, execution, and refinement without human intervention. Through iterative BO, ESA rapidly converged to near-optimal operating conditions, raising the final solution transmittance from 82\% to 99.7\%, which corresponds, via the Beer--Lambert law, to a 98.5\% reduction in residual dye concentration and near-complete decolourisation. This validation shows the framework's capacity for closed-loop, data-driven optimisation in a real laboratory.

In summary, this paper makes the following key contributions:
\begin{itemize}
    \item We frame next-generation self-driving laboratories around \emph{learnability} along three interdependent axes---cognitive, experimental, and embodied---and formalise this perspective as a design principle for embodied scientific intelligence.
    \item We realise this principle as the Embodied Science Agent, coupling digital cognition with physical embodiment through three synergistic modules---the Digital Agent, the Embodied Agent, and the EUI---to enable adaptive, self-evolving discovery in real-world environments.
    \item We demonstrate the framework in a real-world Electro-Fenton degradation experiment for Safranine T wastewater treatment, autonomously planning, executing, and optimising a complex chemical process without human intervention.
\end{itemize}
\begin{figure*}
    \centering
    \includegraphics[width=\linewidth]{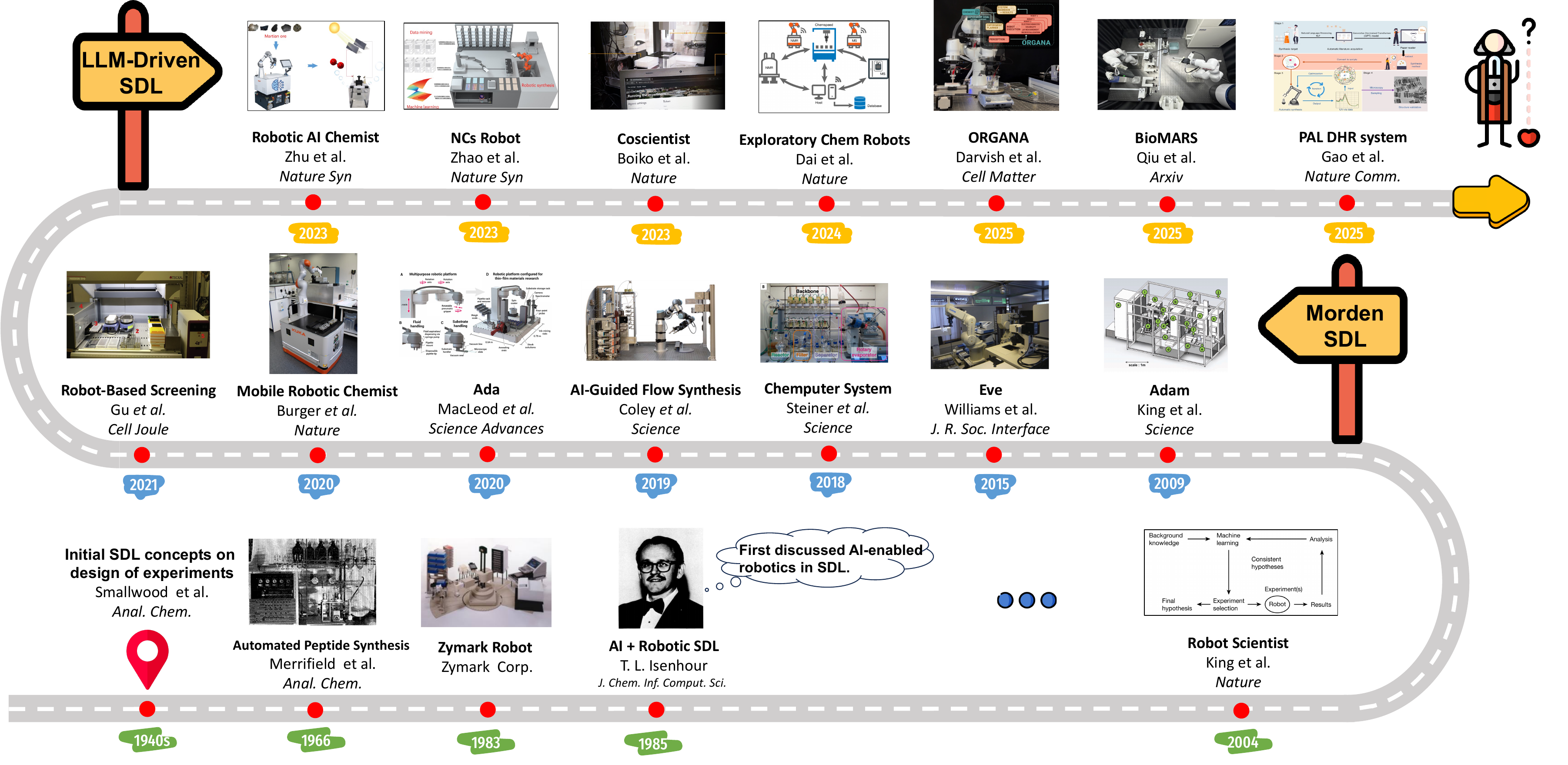}
    \caption{The Evolution of Self-Driving Laboratories. This timeline traces the development of autonomous experimentation systems from initial concepts in the 1940s to foundation model integration in recent years, spanning three eras: \textbf{Early Era} (1940s--2009, bottom) establishing foundational principles from design of experiments to closed-loop robot scientist; \textbf{Modern Era} (2009--2022, middle) deploying specialised robotic platforms across scientific domains; and \textbf{LLMs Era} (2022--present, top) integrating foundation models for high-level reasoning. }
    \label{fig:related_work}
\end{figure*}
\section{Related Work}
\label{sec:related}

Self-driving laboratories, defined as autonomous systems capable of formulating hypotheses, designing experiments, executing protocols, and iterating without human intervention, represent a grand challenge at the intersection of artificial intelligence, robotics, and scientific discovery. Over eight decades, SDLs have evolved from statistical experimental design frameworks through specialised robotic platforms to foundation model-driven systems, progressively expanding the scope of autonomous experimentation (Fig.~\ref{fig:related_work}). Yet this evolution reveals a critical tension: increasing cognitive sophistication has come at the cost of physical grounding.

\subsection{Three Eras of Self-Driving Laboratories}
\label{sec:sdl_overview}

\textbf{Early Era (1940s--2009).}
The conceptual foundations of SDLs were established long before the enabling technology existed. Laboratory automation began in the 1940s with parallel catalyst screening for the Haber-Bosch process, and was formalised by Smallwood's work on the design of experiments~\cite{smallwood1947design}, which introduced the statistical planning principles that remain central to modern SDLs. The 1960s brought Merrifield's automated peptide synthesiser~\cite{Merrifield1965, merrifield1966instrument}, demonstrating reliable automation of complex sequential chemistry, alongside growing attention to lab automation. Closed-loop simplex optimisation of chromatography and reaction conditions followed in the 1970s--80s~\cite{Berridge1982}, while Isenhour's ``Robotics in the Laboratory''~\cite{isenhour1985robotics} presciently articulated AI-driven experiment planning. Combinatorial chemistry and high-throughput screening then established feedback-driven workflows in the 1990s~\cite{Gordon1994}. The 2000s saw a resurgence driven by microfluidic and flow-chemistry platforms for closed-loop synthesis and reaction discovery~\cite{Krishnadasan2007, parrott2011self, Kreutz2010, Adamo2016Science, Christensen2021Science, Mehr2020Science, Granda2018Nature}, and by materials-discovery systems integrating Bayesian optimisation and neural networks for autonomous exploration~\cite{Raccuglia2016Nature, Gongora2020SciAdv, Häse2018, Roch2020, stach2021autonomous}.

\textbf{Modern Era (2009--2022).}
The first genuine robot scientist emerged in 2009 with King et al.'s Adam~\cite{King2009Automation}, which autonomously generated and tested hypotheses about yeast functional genomics through robotic execution, later extended by Eve for drug discovery~\cite{Williams2015}. This era was characterised by increasingly capable, purpose-built robotic platforms: Cronin's modular system driven by a chemical programming language that separates synthesis logic from hardware~\cite{Steiner2019ModularRoboticSynthesis}, Coley et al.'s flow-synthesis platform coupling AI retrosynthetic planning with automated execution~\cite{Coley2019FlowSynthesisAI}, Cooper's mobile robotic chemist operating within a conventional laboratory using computer vision and active learning~\cite{Burger2020MobileRoboticChemist}, self-driving platforms for thin-film and perovskite materials combining robotic deposition, automated characterisation, and Bayesian optimisation~\cite{MacLeod2020SelfDrivingThinFilm, Gu2020AntisolventScreening}, and a portable platform for universal chemical synthesis emphasising accessibility~\cite{Manzano2022PortableSynthesis}.

\textbf{LLM Era (2022--Present).}
The integration of large language models transformed SDL capabilities, enabling systems to leverage vast pre-trained scientific knowledge and orchestrate tools via natural language. Coscientist~\cite{Boiko2023LLMScience} uses GPT-4 to plan experiments and control hardware, ChemCrow~\cite{Bran2024ChemCrow} orchestrates synthesis planning and execution, and AI4Sci Agent~\cite{ding2025scitoolagent} unifies cross-disciplinary API coordination. Recent systems push scope and autonomy further: catalyst synthesis from Martian meteorites~\cite{zhu2024automated}, colloidal nanocrystal synthesis~\cite{Zhao2023ColloidalNanocrystals}, autonomous mobile robots for exploratory chemistry~\cite{dai2024autonomous}, the ORGANA robotic chemistry assistant~\cite{darvish2025organa}, bimanual manipulation for autonomous cell culture~\cite{biomars2025}, and complete end-to-end nanoparticle platforms producing research-ready materials~\cite{Gao2025NanoparticleE2E}. SDLs have correspondingly matured into critical research infrastructure across chemistry, biology, and materials science~\cite{Lunt2024, szymanski2023autonomous, PyzerKnapp2022}. This trend is reinforced by LLM-driven advances in reaction prediction~\cite{schwaller2021, coley2017}, retrosynthesis~\cite{segler2018planning, chen2020retrosynthesis}, materials design~\cite{jablonka2024gpt, merchant2023scaling, kang2024chatmof}, protein engineering~\cite{hie2024efficient, madani2023large}, gene editing~\cite{qu2025crispr}, and clinical applications~\cite{singhal2023large, thirunavukarasu2023large, li-etal-2024-mmedagent, goodell2025large}, with similar momentum in chip design~\cite{wu2024chateda}, mechanical prototyping~\cite{ni2024mechagents}, geospatial analysis~\cite{huang2024geoagent}, neuroscience~\cite{luo2025large}, and hazard response~\cite{xie2025marsha}.

\subsection{Open Challenges: The Embodiment Gap}
\label{sec:open_oppo}
Despite this remarkable progress, current foundation model-era SDLs share a fundamental architectural inadequacy rooted in the absence of embodied intelligence: they treat experimentation as abstract symbol manipulation rather than embodied physical practice. Human scientists, by contrast, develop experimental expertise through embodied interaction, acquiring tacit sensorimotor knowledge---that certain colour changes indicate reaction completion, that reagent viscosity requires adjusted pipetting---that enables adaptive problem-solving when experiments deviate from expectations. Current architectures maintain a strict separation between disembodied foundation models (reasoning and planning) and robotic systems (executing pre-programmed instructions), which manifests in three critical limitations. \textbf{(1) Lack of physical grounding:} plans are generated from statistical patterns in the literature rather than causal models grounded in physical experience, yielding instructions that are syntactically correct but physically infeasible (e.g., stirring speeds that cause splashing, heating rates that decompose intermediates), and the systems lack the perceptual grounding to read real-time visual indicators. \textbf{(2) Rigid separation preventing adaptive execution:} the planning module cannot perceive unexpected conditions and the execution module cannot reason about them, so deviations such as spills or equipment malfunctions typically cause catastrophic failure or require human intervention. \textbf{(3) Absence of sensorimotor learning:} while modern SDLs optimise experimental parameters through active learning, they do not improve their fundamental motor skills or perceptual judgements through practice---a system executing its thousandth titration performs no better than its first.

These limitations share a root cause: treating language models as disembodied planning oracles that emit instructions for separate robotic executors contradicts how human experimental competence emerges, through the inseparable integration of conceptual understanding, perceptual skill, and motor competence. Addressing this gap requires a paradigm shift toward \textit{embodied science agents} that unify perception, cognition, and action, developing grounded understanding through direct sensorimotor interaction. Section~\ref{sec:method} introduces the Embodied Science Agent architecture that instantiates these principles.

\section{Methodology}
\label{sec:method}

\begin{figure*}
    \centering
    \includegraphics[width=\linewidth]{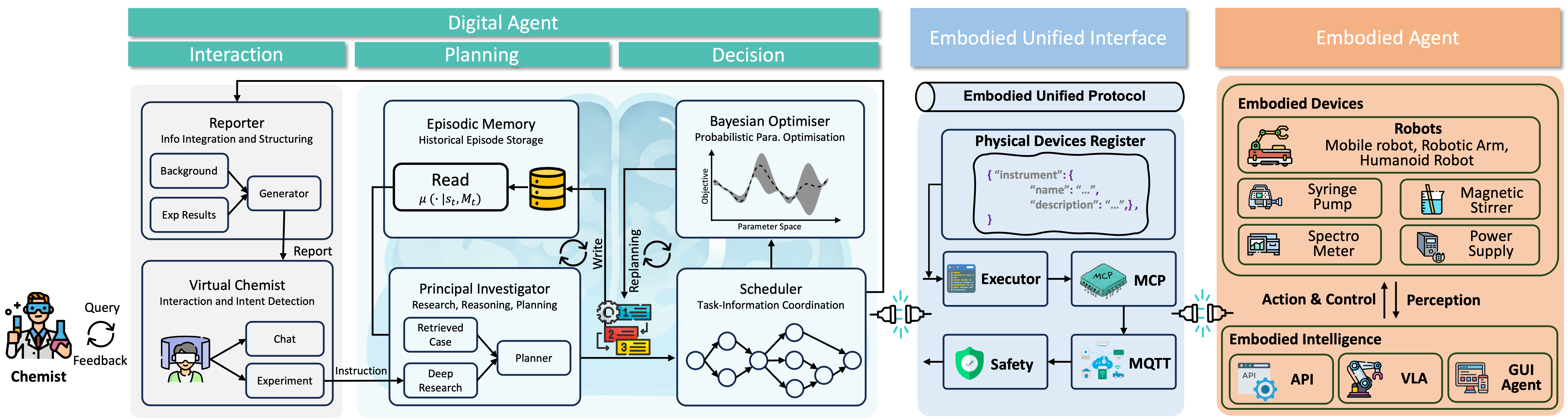}
    \caption{The architecture of the Embodied Science Agent (ESA), a hybrid
    cognitive--embodied paradigm for self-driving laboratories with three
    components. The \textbf{Digital Agent}, the cognitive core and central
    novelty of ESA, evolves open-endedly by integrating online knowledge,
    embodied experimentation, and past experience for reasoning, planning,
    and decision-making. The \textbf{Embodied Agent} orchestrates diverse
    robots and scientific instruments through multiple control modules for
    generalised, reliable, and adaptive operation. The \textbf{Embodied
    Unified Interface} connects digital intelligence with heterogeneous
    laboratory devices via standardised, low-latency, and safe protocols.
    Together, they close the loop between perception, reasoning, and
    embodied execution for autonomous scientific discovery.}
    \label{fig:overview_ESA}
\end{figure*}

As shown in Fig.~\ref{fig:overview_ESA}, ESA is built on a hybrid cognitive--embodied architecture that tightly couples digital intelligence with physical experimentation. It integrates a cognitive reasoning core, the \textbf{Digital Agent} (analogous to the cerebrum), which handles hypothesis generation, planning, and memory-based inference, with embodied experimental systems, the \textbf{Embodied Agent} (analogous to the cerebellum), which translates plans into accurate physical execution; the \textbf{EUI} bridges the two so that abstract reasoning is grounded in reliable embodied action. We next present an overview of the workflow (Section~\ref{sec:overview_ESA}) followed by its core components: open-ended evolution (Section~\ref{sec:open_evo}), the EUI (Section~\ref{sec:EUI}), and the Embodied Agent (Section~\ref{sec:embodied_agent}).

\subsection{Overview of Embodied Science Agent}
\label{sec:overview_ESA}
For a specific experiment, the workflow is an iterative cycle coupling digital decision-making with embodied execution. It begins with the \textit{Virtual Scientist Avatar}, a multimodal (ASR/TTS) interface for transparent human--agent interaction and monitoring. When an experimental intent is identified, it is passed to the \textit{Principal Investigator (PI) Agent} in the \textbf{Digital Agent} layer (the cerebrum), which performs deep research for external knowledge and retrieves relevant past cases from its episodic memory, shifting planning from static rules to adaptive, experience-driven reasoning.

The resulting graph-based plan is dispatched to the \textbf{EUI} (the \textit{Executor} in Fig.~\ref{fig:overview_ESA}), which harmonises data and control via the Model Context Protocol (MCP), a device registry, and embedded safety checks, and drives the \textbf{Embodied Agent} to execute the plan through robotic systems and instruments. After execution, the Embodied Agent returns logs and objectives to the Scheduler, which forwards the objectives to the \textit{Bayesian Optimiser} to propose the next configurations; these are replanned by the PI Agent and the cycle repeats until termination, whereupon the \textit{Reporter Agent} compiles a final report that the Virtual Scientist presents to the user.

\subsection{Open-Ended Optimisation and Evolution}
\label{sec:open_evo}
Open-ended evolution is the core self-improvement capability of ESA. Whereas conventional SDLs optimise within task-specific, manually designed objectives, ESA self-improves across tasks by integrating prior knowledge with new experience---crucial for discovery settings where target phenomena or success criteria are not known in advance. It is realised through two complementary mechanisms: (1)~a Learning-from-Memory mechanism that retrieves, generalises, and reuses past knowledge to guide reasoning, and (2)~BO that drives data-efficient exploration of the parameter space. Together they form a closed adaptive loop advancing from static automation toward self-improving scientific discovery.

\subsubsection{Learning-from-Memory Mechanism}
\label{sec:memory}
To support reliable planning, the PI Agent recalls and reuses historical experiences rather than reasoning in isolation. Analogous to the hippocampus, episodic memory complements the language model's abstract deliberation by grounding decisions in concrete experimental evidence. We implement it as a case bank $M_t$ operated through two atomic functions, Write and Read. Each case is represented as a tuple $(s_t,a_t,r_t)$, where the state $s_t$ encodes the experimental query together with contextual background derived from deep research, the action $a_t$ specifies the planned experimental procedures, and the outcome $r_t$ records the measured results such as yield, spectrum, or efficiency. The Write function appends each newly executed experiment into the memory,
\begin{equation}
\text{Write}(s_t,a_t,r_t,M_t) = M_{t+1} = M_t \cup {(s_t,a_t,r_t)},
\end{equation}
ensuring that both successful and failed trajectories are retained. The Read function retrieves relevant past cases by computing the similarity between the current query state and stored states.
\begin{equation}
\text{Read}(s_t,M_t) = \mathop{\mathrm{TopK}}_{(s_i,a_i,r_i)\in M_t} \operatorname{sim}(\operatorname{enc}(s_t), \operatorname{enc}(s_i)),
\end{equation}
where $\operatorname{enc}(\cdot)$ is a pretrained encoder and $\operatorname{sim}(\cdot)$ denotes cosine similarity. This retrieval process yields the most relevant past cases, returning both their procedures and outcomes as candidate guidance for planning new experiments.

Rather than merely storing past results, the memory acts as a cognitive substrate for continual evolution: negative outcomes delimit unproductive regions of the search space while positive trajectories become structured priors, so that as the episodic memory grows the decision system progressively refines its planning strategies and causal understanding.

\subsubsection{Bayesian Optimisation}
\label{sec:bo}
BO globally optimises a black-box objective $f(x): \mathcal{X} \mapsto \mathbb{R}$ whose form is unknown and whose evaluation is expensive. BO treats $f$ as a \textit{random function} and maintains a probabilistic \textit{surrogate} $\hat{f}(x|\mathcal{D})$ of its posterior given observed data $\mathcal{D}$; an \textit{acquisition function} scores candidates from this surrogate, and its maximiser is chosen as the next query point. It proceeds iteratively: the surrogate is fit to the current dataset, the acquisition function is optimised to propose the next configuration, the black-box is evaluated, and the new observation is appended, until a stopping criterion is met. The full anatomy of BO, an illustration of how it integrates with ESA, and a practical instantiation via Gaussian processes and the upper confidence bound (GP-UCB) are given in Appendix~\ref{app:gp-ucb}, with pseudocode in Appendix~\ref{app:bo_pseudo_code}.

Within the ESA framework, BO optimises the experiment configuration \( x \) by maximising the expected utility \(\mu(\pi_{wf}, x)\) over both the workflow policy \(\pi_{wf}\) and the configuration \( x \), either for a fixed \(\pi_{wf}\) or as a nested component in a hierarchical scheme that jointly optimises the two. BO is well-suited to this task: its probabilistic surrogate quantifies uncertainty and its acquisition function balances exploration and exploitation for high sample efficiency, while its global-search design and robustness to noise suit the expensive, stochastic utility \(\mu(\pi_{wf}, x)\), which lacks a closed form and requires running the full workflow to evaluate.

\subsection{Embodied Unified Interface}
\label{sec:EUI}
Modern laboratories host diverse instruments from different manufacturers, operating on incompatible protocols and control latencies, which makes integrating LLM-driven digital agents with embodied systems a major challenge. To bridge this gap, the EUI consolidates parallel orchestration, tool invocation, safety validation, and embodied execution into a single verifiable contract, grounding high-level reasoning in self-driving experimentation.

Unlike existing protocols such as SiLA 2~\cite{juchli2022sila}, or HELAO~\cite{rahmanian2022enabling}, which address device interoperability or workflow description in isolation, EUI provides a contract-based mechanism for transforming reasoning into execution: intents are structured through large-model function calling and validated against formal schemas; a DAG-based scheduler decomposes them into a transactional workflow with dependency checking and parallel execution; each subtask is bound to an MCP-defined tool contract; and verified contracts are compiled into MQTT messages for reliable, auditable communication. Safety guards are embedded throughout---schema validation, physical-constraint checking, timeout fallback, emergency-stop channels, and sandboxed execution---so that reasoning, orchestration, and execution share one verifiable framework.

\subsection{Embodied Agent}
\label{sec:embodied_agent}

A central capability of ESA is its operation across heterogeneous physical devices, moving beyond the narrow emphasis on robotic manipulation in prior work. We propose a unified control paradigm that categorises devices by the form of embodied intelligence their control requires, spanning the continuum from digital parameter setting to embodied physical manipulation: programmable devices with APIs, interface-bound devices with GUIs, and physical apparatus requiring direct manipulation via vision-language-action (VLA) models. To further enhance the reliability and versatility of embodied experimentation, we also design a new scientific robotic platform, detailed in Section~\ref{sec:rsr-robot}.

\subsubsection{API-based Control for Programmable Devices}

Instruments exposing programmable APIs are operated directly through digital signals for machine-verifiable, reproducible control. Formally, each device exposes callable functions $\mathcal{F} = \{f_1, \dots, f_M\}$ (e.g., start, stop, set parameter, read value); a high-level instruction $l_t$ is mapped to a sequence of API calls $\mathbf{c}_t = (f_{i_1}(\theta_1), f_{i_2}(\theta_2), \dots)$ whose execution updates the device state and returns observations $o_t$ to the decision pipeline. This gives structured, deterministic, and traceable control without any reinforcement-learning or MDP formulation.

\subsubsection{GUI-based Control via Interface-driven Devices}

Devices lacking APIs but exposing graphical user interfaces are driven by a GUI agent that issues mouse, keyboard, or touch events. At each step it receives multimodal observations $o_t = \{I_t, s_t, l_t\}$ (screen image, internal state such as cursor position, and language goal) and produces atomic UI actions $a_t = \{a_t^1, \dots, a_t^L\}$ (clicks, key presses, text input). We model this as a reward-free Markov decision process $\mathcal{M}_\text{GUI} = (\mathcal{O}, \mathcal{A}, P, \rho_0, \gamma)$ trained via imitation learning on expert demonstrations $\mathcal{D} = \{(o_t, a_t)\}_{t=1}^T$. In practice, pH meters and electrochemical workstations are operated in a closed-loop ``see--think--act--evaluate'' cycle: an Orchestrator generates atomic action plans and the agent parses visual cues to verify outcomes and adapt subsequent actions, enabling reliable end-to-end operation of non-programmable instruments.

\subsubsection{Robotic Manipulation with VLA}
\label{sec:vla_modelling}
Physical apparatus such as test tubes and beakers cannot be operated through digital interfaces and require direct embodied manipulation. For this, ESA employs VLA-based robotic policies that couple language-driven reasoning with visual grounding and continuous motor control, translating temporally grounded textual plans into fine-grained actions and generalising across novel tools, materials, and layouts while maintaining safety and reproducibility.

Motivated by these principles, we first formalise the scientific laboratory manipulation problem as a language-triggered selection among multiple visuomotor policies, followed by the generation of corresponding action sequences. We implement a Visuomotor Policy Dictionary (VPD), a keyed collection of language-conditioned visuomotor subpolicies. Retrieval and execution of these subpolicies enable coordinated bimanual manipulation. At each time step $t$, the robot receives multimodal observations $o_t = \{I_t, s_t, l_t\}$, where $I_t \in \mathbb{R}^{N \times H \times W \times 3}$ denotes RGB images with $H\times W$ resolution captured by $N$ cameras, $s_t$ is the embodiment's proprioceptive state, and $l_t$ is a language instruction encoding the task goal. From these observations, the robot generates a sequence of actions $a_t \in \mathbb{R}^{L \times d}$ corresponding to joint-angle targets over a horizon of $L$ steps. As in the GUI-based control setting above, we cast the problem as a reward-free MDP $\mathcal{M} = (\mathcal{O}, \mathcal{A}, P, \rho_0, \gamma)$ and train the embodiment via imitation learning on expert demonstrations $\mathcal{D} = \{(o_t, a_t)\}_{t=1}^{T}$.

Multiple visuomotor policies are trained, each associated with a semantic label (e.g., ``insert electrode,'' ``transfer beaker''). Each label-policy pair forms an entry in the VPD, enabling language-conditioned retrieval and continuous bimanual action generation. Following \citet{zhao2023learningfinegrainedbimanualmanipulation}, policy networks are implemented using an Action Chunking Transformer (ACT), which predicts distributions over action chunks conditioned on visual and proprioceptive sequences. Each semantic label corresponds to a subpolicy $\pi_\theta^{(k)}$, and all subpolicies are jointly optimised via multi-task training:
\[
\min_{\{\theta_k\}} \sum_{k=1}^{K} \mathbb{E}_{(o_t,a_t) \sim \mathcal{D}_k} \big[ \mathcal{L}(a_t, \pi_{\theta_k}(o_t)) \big],
\]
where $K$ is the number of subpolicies and $\mathcal{D}_k$ is the subset of demonstrations for the $k$-th label. At runtime, the embodiment selects the most appropriate visuomotor policy via language matching. This VPD-based framework preserves action continuity, safety, and high precision, meeting the strict requirements of automated chemical experimentation.



\subsection{Bimanual Scientific Robot}\label{sec:rsr-robot}
As discussed in Section~\ref{sec:vla_modelling}, many scientific instruments and laboratory operations still rely on direct physical handling, such as pouring, pipetting, or connecting apparatus, rather than programmable or GUI-based control. Unlike digital interfaces with relatively standardised workflows, these physics manipulation tasks require a robot capable of precise manipulation in complex laboratory environments, with strong generalisation across diverse tools, procedures and environments. To address these challenges, we designed and developed a bimanual mobile manipulator that integrates visual perception, modular end-effector adapters and dual-arm manipulation to interact safely and flexibly with a wide range of experimental instruments. As the physical embodiment of ESA, it bridges high-level digital reasoning with real-world scientific experimentation.
\subsubsection{Overview of the Robotic Platform}
\label{sec:rsr-robot-implementation}
\begin{figure}[!t]
\centering
\includegraphics[width=\linewidth]{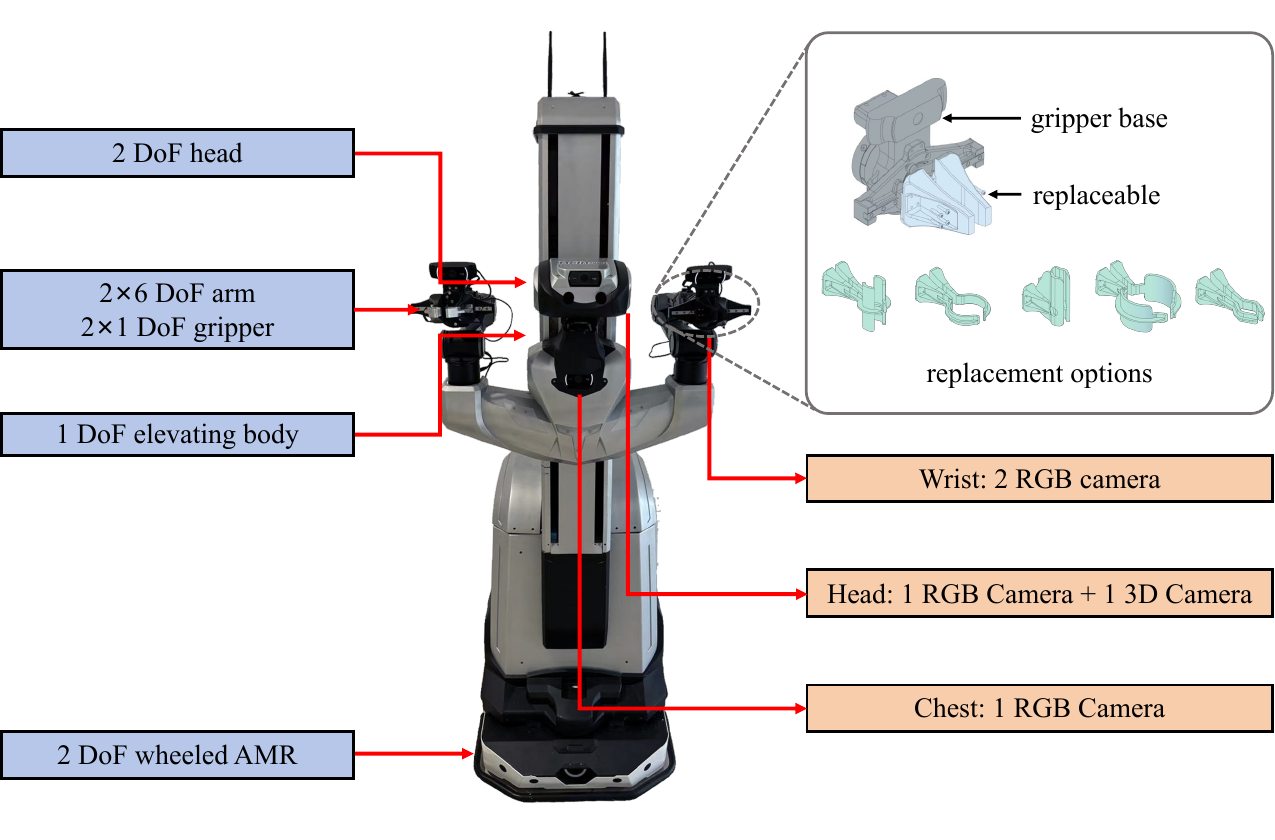}
\caption{Overview of the robotic platform.
The dual-arm robot comprises two 6-DoF robotic arms equipped with 1-DoF grippers, a 2-DoF articulated head, an omnidirectional AMR base offering 2 planar DoFs for holonomic motion, and a 1-DoF lift for vertical adjustment. These degrees of freedom are highlighted in the blue box. The platform also integrates a synchronised multi-camera sensing suite consisting of one RGB and one depth camera mounted on the head, two RGB cameras on each wrist, and one RGB camera on the chest, as indicated in the orange box. Together, these components provide rich proprioceptive and multi-view visual feedback to support both global perception and fine-grained manipulation.}
\label{fig:rsr_show}
\end{figure}

As shown in Fig.~\ref{fig:rsr_show}, the dual-arm robot is a general-purpose robotic platform designed for automated chemical experimentation. It is equipped with two 7-degree-of-freedom (DoF) arms, each consisting of six rotational joints and one prismatic joint that supports interchangeable end-effectors, enabling versatile manipulation across a wide range of laboratory tasks.
Its omnidirectional Autonomous Mobile Robot~(AMR) chassis provides robust mobility with three planar degrees of freedom for translation and yaw rotation, while an integrated lift mechanism adds a vertical degree of freedom to adjust the manipulator's working height. By integrating these components, the robot can seamlessly navigate and interact with vertically diverse laboratory setups, ensuring stable and efficient operation in dynamic experimental environments.

For perception, the robot combines a 31-dimensional proprioceptive state (dual-arm joint positions and velocities, gripper displacements, head pose, lift height, and chassis motion) with five synchronised camera streams (a head-mounted depth sensor, a forward-facing head camera, per-wrist cameras, and a chest camera), fusing holistic scene understanding with local perception for accurate contact localisation. To handle laboratory objects of varying shape, size, and fragility that conventional two-finger grippers grasp unreliably, we further equip it with a suite of modular gripper adapters mounted via a standardised quick-change interface, each optimised for a category of instruments (e.g., cylindrical tubes, glassware, electrodes); details are given in Section~\ref{sec:gripper-adapter-suite}. The dual-arm system runs at a 30~Hz control frequency with hardware-level PD controllers issuing parallelised commands to the arms, grippers, head, lift, and chassis, complemented by emergency-stop circuitry and continuous workspace monitoring for safe operation.

\subsubsection{Modular End-Effector Adapters for Laboratory Manipulation}
\label{sec:gripper-adapter-suite}
Motivated by the limitations of existing end-effectors for robotic chemistry automation, we propose a modular tooling approach that materially increases the robot's adaptability across heterogeneous laboratory instruments. 
Most commercial grippers are engineered for regular, rigid workpieces. Laboratory vessels such as reagent bottles, pipettes, beakers, electrodes, and graduated cylinders, however, exhibit large variability in shape, size, and material, and are often made of fragile, low-friction glass or plastic.
This diversity imposes conflicting requirements on grasp geometry and contact-force regulation: a firm grip risks damaging fragile items, while a compliant grip may yield unstable contacts and slippage. Consequently, a single, monolithic gripper often cannot switch reliably between distinct classes of laboratory objects without mechanical reconfiguration or fixture replacement, which increases setup time and undermines experimental repeatability.

To mitigate these challenges, we developed a gripper adapter suite: a modular family of quick-change add-on components that augment a baseline two-finger commercial gripper, as shown in Fig. \ref{fig:rsr_show}. The design preserves the gripper's original kinematics, actuation and control interface while introducing a standardised distal quick-change interface that permits rapid installation and removal of interchangeable adapter modules. Each adapter is tailored to a target class of laboratory items (e.g., cylindrical tubes, electrodes, beakers), with geometry and contact surfaces optimised to provide compliant, distributed contact and controlled force transfer. By combining multiple specialised fingertip geometries with fast physical reconfiguration, the adapter suite enables flexible, task-appropriate grasps without requiring reprogramming or complex mechanical adjustments.

In practice, the gripper adapter suite yields three principal benefits for automated laboratory workflows: (1) improved grasp stability through geometry-specific contact design, (2) enhanced compliance and reduced risk of damage for fragile or low-friction materials, and (3) faster experimental setup and higher repeatability by removing the need for manual fixture swaps or controller retuning. Implementation and additional configuration details of the gripper adapter suite are provided in the Appendix \ref{app:gripper-adapter-suite-details}.

\section{Experiment}
Industrialisation discharges large volumes of dye-containing wastewater, and conventional methods such as adsorption or biological degradation are ineffective against highly stable organic dyes, merely transferring pollutants rather than breaking them down. The Fenton reaction, a leading advanced oxidation process, generates strongly oxidative hydroxyl radicals ($\cdot$OH) but works optimally only under acidic conditions and produces large amounts of iron sludge. The Electro-Fenton (EF) process addresses this by electrochemically generating H$_2$O$_2$ and continuously regenerating Fe$^{2+}$/Fe$^{3+}$ on the electrode, while reusing waste Fenton sludge for composite electrode fabrication enables sustainable solid-waste recycling. Since optimising such multi-parameter electrochemical systems manually is slow and inefficient, we adopt the Electro-Fenton degradation of Safranine T as a representative scenario for deploying ESA as a self-driving laboratory that autonomously executes and optimises the process.

The remaining sections are organised as follows: Section~\ref{sec:exp_param} introduces the parameter space; Section~\ref{sec:exp_lab} details the implementation of the embodied self-driving laboratory; Section~\ref{sec:exp_T} reports the results of the Safranine T degradation experiments; finally, Section~\ref{sec:vla} presents an embodied manipulation case study that demonstrates the agent's generalisation and reasoning capabilities in real-world chemical experiments.

\subsection{Parameter Optimisation}
\label{sec:exp_param}

We use Safranine T as the model pollutant to evaluate the Electro-Fenton degradation performance of the fabricated iron-sludge composite electrode. Based on their critical influence on hydroxyl radical generation and pollutant degradation, we optimise five key operational parameters:
\begin{itemize}
    \item \textbf{pH:} controls the Fe$^{2+}$/Fe$^{3+}$ cycle and the stability of hydroxyl radicals.  
    \item \textbf{Voltage:} determines the potential applied across the electrodes, affecting H$_2$O$_2$ generation and the rate of radical formation.
    \item \textbf{Aeration rate:} determines the supply of O$_2$, which influences H$_2$O$_2$ generation.  
    \item \textbf{Stirring rate:} governs mass transfer and the distribution of reactive species within the solution. 
    \item \textbf{Na$_2$SO$_4$ concentration:} acts as the supporting electrolyte, affecting conductivity and the electrochemical environment.  
\end{itemize}



Crucially, these parameters do not act in isolation but exhibit strong interaction effects that create a complex optimisation landscape: higher voltage raises the degradation rate but at low pH accelerates electrode corrosion and gas evolution, while pH also governs iron speciation (precipitation at high pH, suppressed radical generation at very low pH) and high Na$_2$SO$_4$ can promote parasitic side reactions. The resulting search space renders conventional one-factor-at-a-time or manual tuning impractical, motivating automated optimisation.

Within the ESA framework, a BO module guides parameter tuning, efficiently exploring the high-dimensional experimental space to identify globally optimal operating conditions that maximise the final solution transmittance (ranging from $0$\% to $100$\%). The discretised search space is summarised in Table~\ref{tab:param_ranges}. At the instrument resolution used for embodied execution, these five parameters yield over $2\times10^{8}$ unique experimental configurations, far beyond the reach of manual or one-factor-at-a-time tuning.

\begin{table}[t]
\centering
\caption{The discretised search space for Electro-Fenton parameter optimisation.}
\label{tab:param_ranges}
\resizebox{\linewidth}{!}{
\begin{tabular}{lcccc}
\toprule
\textbf{Parameter} & \textbf{Symbol} & \textbf{Range / Set} & \textbf{Step Size} & \textbf{Unit} \\
\midrule
Acidity & pH & $\{2.0, 2.5, 3.0, 4.0, 5.0\}$ & (Set) & -- \\
Applied Voltage & $U$ & $1.00$--$10.00$ & $0.01$ & \si{\volt} \\
Aeration Intensity & PWM & $50$--$255$ & $1$ & -- \\
Stirring Rate & $\omega$ & $100$--$300$ & $10$ & \si{rpm} \\
Na$_2$SO$_4$ Conc. & $C_{\text{Na$_2$SO$_4$}}$ & $25$--$75$ & $5$ & \si{\milli\mol\per\liter} \\
\bottomrule
\end{tabular}
}
\end{table}

\subsection{Embodied Self-Driving Laboratory}
\label{sec:exp_lab}
To automate electrochemical wastewater treatment, we constructed an embodied self-driving laboratory that integrates a heterogeneous suite of robotic and analytical instruments and autonomously executes a closed loop of experimentation, measurement, and parameter optimisation, controlled by the unified embodied-agent framework of Section~\ref{sec:embodied_agent}.

The three device categories of Section~\ref{sec:embodied_agent} are all instantiated here; full device specifications, the complete instrument list, and their exposed actions are given in Appendix~\ref{app:experimental-devices-details} (Fig.~\ref{fig:exp_devices} and Table~\ref{tab:devices-actions}). \textbf{API-based} devices are wrapped through a unified layer that abstracts their native Serial/Modbus protocols, each implemented as an agent inheriting a common \texttt{BaseDevice} class with standardised initialisation, state tracking, and error handling. \textbf{GUI-based} instruments lacking APIs are driven by vision-based agents using OmniParser~\cite{lu2024omniparserpurevisionbased} in a closed-loop ``see--think--act--evaluate'' cycle that reads on-screen values and navigates the software to configure, monitor, and export data. \textbf{Physical} apparatus requiring direct manipulation is handled by two collaborative arms under the ACT framework and a centralised, collision-free motion planner: one arm positions and immerses the pH probe while the other retrieves, cleans, and places the electrode.

In a typical run (\SI{25}{\celsius}), the system prepares the reaction mixture ($20\,\text{mg L}^{-1}$ Safranine T, $100\,\text{mmol L}^{-1}$ Na$_2$SO$_4$, deionised water), acidifies it to the target pH with $0.2\,\text{mol L}^{-1}$ H$_2$SO$_4$ under robotic pH monitoring, and performs a 30-min pre-treatment (with the second arm cleaning the electrodes) followed by a 60-min electrochemical treatment at the specified conditions. The final transmittance is then measured and submitted to the Bayesian optimiser, which proposes the next configuration before a cleaning protocol resets the apparatus.

\begin{figure}[h]
\centering
\includegraphics[width=\linewidth]{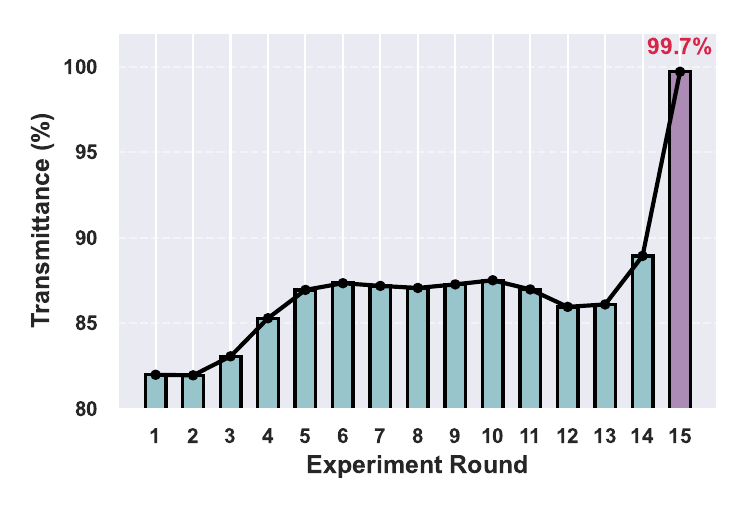}
\caption{Transmittance results across 15 experimental rounds.}
\label{fig:exp_result}
\end{figure}

\subsection{Experimental Results}
\label{sec:exp_T}

To evaluate the system's autonomous optimisation capability, we conducted a Safranine T degradation experiment and monitored the evolution of transmittance. To initialise the optimiser, the system first performs 5 rounds of Latin hypercube sampling to ensure representative coverage of the parameter space before proceeding with autonomous iterative optimisation. This was followed by 10 autonomous experimental iterations guided by the Bayesian optimiser. Figure~\ref{fig:exp_result} presents the complete trajectory across all 15 experimental rounds, revealing a clear progression through three distinct phases.

In the \textbf{initialisation phase} (rounds 1--5), the five LHS data points trained the initial Gaussian Process surrogate, with transmittance improving gradually from a baseline of $\sim$82\% to $\sim$87\%. The \textbf{plateau phase} (rounds 6--13) showed stable values fluctuating between 86\% and 88\%, reflecting the optimiser's exploration--exploitation trade-off as it refined local optima while probing less-explored regions. In the \textbf{breakthrough phase} (rounds 14--15), transmittance rose to 89\% in round 14 and then leapt to 99.7\% in the final round, a 17.7-percentage-point gain over baseline. Because transmittance relates to residual dye concentration through the Beer--Lambert law ($A = -\log_{10} T \propto c$), this corresponds to a drop in residual absorbance from $\approx$0.086 to $\approx$0.001, i.e., a 98.5\% reduction in residual Safranine T concentration and near-complete decolourisation. We note that spectrophotometric decolourisation characterises chromophore removal and does not by itself establish full mineralisation.

The efficiency gains are substantial. A traditional manual campaign over a search space of this scale, with sufficient replicates for statistical significance, would demand roughly two researchers for about 30 days; our prototype reached this expert-level configuration on a single set of equipment within roughly 45 hours of continuous autonomous operation.

Crucially, this rapid convergence is not attributable to the optimiser alone, but to the synergy between the three layers of ESA. The Embodied Agent's adaptive manipulation ensured precise, repeatable handling and stable sensing, minimising physical execution noise and thus supplying the high-fidelity feedback that a reliable surrogate model requires; the Digital Agent's memory-derived priors constrained the search to chemically plausible regions, preventing wasted iterations. It is on this foundation of low-noise measurements and valid search bounds that the Bayesian optimiser could efficiently escape local optima: the autonomous discovery observed here is an emergent property of coupling cognitive reasoning, experimental optimisation, and embodied execution, not of any single component alone.





\subsection{Embodied Manipulation Case-study}
\label{sec:vla}
We demonstrate the robot's embodied learnability through a representative case study, showing that high-level textual plans generated by ESA are reliably executed through specialised visuomotor policies and remain robust to variation in object placement. The task is a simple command---the left arm moves the electrode rack beside the magnetic stirrer, and the right arm inserts the electrode into the beaker---but it stresses the generalisation of the learned Visuomotor Policies Dictionary (VPD): the rack may be placed anywhere within a workspace range, unlike conventional experiments with fixed instrument positions.

As shown in Fig.~\ref{fig:case1_gene_show}, across three sampled rack placements the left arm consistently grasps the rack's edge and moves it towards the stirrer, while the right arm aligns and inserts the electrode. These trials confirm that the VPD translates multi-step textual instructions into precise, stable trajectories, demonstrating strong spatial generalisation in chemical manipulation.

\section{Conclusion}
Guided by a vision of embodied scientific intelligence spanning cognitive, experimental, and embodied learnability, we proposed the Embodied Science Agent (ESA) that integrates digital cognition with physical embodiment through a closed loop of memory-based reasoning, embodied execution, and BO. Validated on the Electro-Fenton degradation of Safranine T with conventional instruments and a dual-arm platform, ESA autonomously planned, executed, and refined experiments without human intervention, raising the final solution transmittance from $82\%$ to $99.7\%$ over 15 rounds---a $98.5\%$ reduction in residual dye concentration under the Beer--Lambert law---while its learned visuomotor policies generalised across varying instrument placements.

These results suggest that autonomous scientific proficiency emerges from the tight coupling of cognition, experimentation, and embodiment rather than any single component. Broader evaluation across chemistries, quantitative baselines against alternative optimisers, and assessment of memory-driven evolution across tasks remain important future work.


\bibliographystyle{ACM-Reference-Format}
\bibliography{reference}

\clearpage
\appendix
\raggedbottom
This appendix provides comprehensive technical details and implementation specifications that support the methodology presented in the main text.
Appendix~\ref{app:learnability} first formalises the notion of learnable experimental design that underpins ESA and maps its three facets onto the system's components.
Appendix~\ref{app:gp-ucb} and Appendix~\ref{app:bo_pseudo_code} then detail the Bayesian optimisation framework, including the mathematical formulation of GP-UCB and the algorithmic pseudocode for autonomous experimental parameter optimisation.
Appendix~\ref{app:vdp-demonstration-dataset-collection} and Appendix~\ref{app:vdp-training-details} describe the construction of the demonstration dataset and the training procedure for the visuomotor policy dictionary, which enables the embodiment to perform diverse manipulation tasks in chemical laboratory settings.
Appendix~\ref{app:gripper-adapter-suite-details} presents the design specifications of the gripper adapter suite that facilitates versatile manipulation of laboratory instruments without hardware reconfiguration.
Finally, Appendix~\ref{app:experimental-devices-details} documents the specifications, control interfaces, and programmable actions of all experimental devices integrated into our automated laboratory platform, forming the foundation of the embodied unified interface layer.

\section{A Formal Perspective on Learnable Experimental Design}
\label{app:learnability}
We formalise the notion of \emph{learnability} that gives the paper its name and connect it to the three components of ESA. Under the hypothetico-deductive method standard in modern laboratories, a scientific task maps a user query and laboratory context to a tuple
\begin{equation}
\text{query} \;\mapsto\; (\mathcal{H}, \mathcal{X}, \mathcal{Y}, f),
\end{equation}
where $\mathcal{H}$ is the hypothesis space of unknown components requiring empirical evidence, $\mathcal{X}$ is the space of experimental designs, $\mathcal{Y}$ the space of observable outcomes, and $f:\mathcal{X}\to\mathcal{Y}$ the (noisy, black-box) experimental procedure that tests those hypotheses. A system is \emph{learnable} to the extent that experience accumulated during operation improves its command of these elements, rather than being frozen at deployment. We distinguish three interdependent facets.

\textbf{Cognitive learnability} concerns forming $(\mathcal{H}, \mathcal{X}, \mathcal{Y}, f)$ from a query. Contemporary SDLs rely on static templates or hard-coded prompts; ESA instead maintains a case bank $\mathcal{M}=\{(q_i,\mathbf{a}_i,r_i)\}_{i=1}^m$ of past queries $q_i$, produced plans $\mathbf{a}_i=(\mathcal{H}_i,\mathcal{X}_i,\mathcal{Y}_i,f_i)$, and outcomes $r_i$, and learns a retrieval policy $\pi(\mathbf{a}\mid q,\mathcal{M})$ that refines its methodological intuition over time. This facet is realised by the \textbf{Digital Agent} (Section~\ref{sec:open_evo}).

\textbf{Experimental learnability} concerns choosing designs $x\in\mathcal{X}$ sequentially so that each is informed by prior observations. Because $\mathcal{H}$ is typically enormous and $f$ is noisy, ESA adopts a Bayesian view of the scientific method: it maintains a belief over outcomes and updates it as evidence arrives, using Bayesian optimisation to navigate to global optima with far fewer experiments than brute-force enumeration. This facet is realised by the \textbf{Bayesian Optimiser} (Section~\ref{sec:bo}, Appendix~\ref{app:gp-ucb}).

\textbf{Embodied learnability} concerns executing $f$ reliably despite mechanical tolerances and environmental noise. Rigid control programs are brittle when spatial configurations change; ESA instead acquires and refines visuomotor skills through demonstration, so that execution remains robust as apparatus is displaced. This facet is realised by the \textbf{Embodied Agent} (Section~\ref{sec:embodied_agent}).

These facets are not independent. Following a Cannikin (weakest-stave) argument, an autonomous scientist is only as capable as its least learnable component: brittle execution or a poor search policy bottlenecks even the strongest reasoning engine. Conversely, they reinforce one another: sharper cognitive priors accelerate the optimiser's convergence, more precise embodied execution lowers the observation noise on $\mathcal{Y}$ and thereby sharpens the surrogate model, and the high-quality data so obtained enriches the memory that the cognitive layer learns from. Autonomous discovery therefore emerges from the co-evolution of the three facets, which is exactly the closed loop instantiated by ESA.

\section{The GP-UCB Implementation of BO}
\label{app:gp-ucb}

\begin{figure}[tb]
\centering
\includegraphics[width=\linewidth]{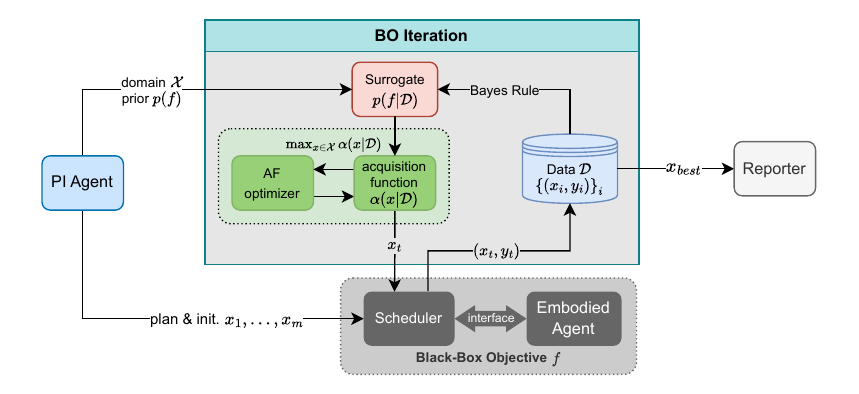}
\caption{Illustration of the Bayesian Optimisation framework as a central component of the ESA architecture. BO receives initial experimental configurations and objectives from the Principal Investigator (PI) Agent. As experiments are conducted by the Embodied Agent, the outcomes are relayed back to BO, providing essential data for refining its surrogate model. BO utilises this feedback to update its predictions and identify the next optimal configuration for experimentation. The Scheduler facilitates the execution of these configurations by coordinating between BO and the Embodied Agent. Ultimately, the best decision obtained during BO is recommended to the Reporter.}
\label{fig:bo_framework}
\end{figure}

A BO algorithm is instantiated by specifying the following components:
\begin{itemize}
\item An \textbf{observation model} $\operatorname{obs}$ that formulates how each observation $y_t = \operatorname{obs}(x_t)$ is correlated to the black-box objective $f(x_t)$, for example through an additive Gaussian noise $y_t = f(x_t) + \epsilon_t$.
\item An \textbf{initialisation strategy} to choose points $x_1, \ldots, x_m \in \mathcal{X}$, thereby obtaining $m$ initial observations.
\item A \textbf{surrogate model} $\hat{f}_{\mathcal{D}}$ to approximate the posterior distribution $p(f|\mathcal{D})$, where $\mathcal{D}$ is the set of historical observations ($x$-$y$ pairs). Usually $p(f|\mathcal{D})$ is inferred from a prior $p(f)$ using Bayes' rule.
\item An \textbf{acquisition function (AF)} $\alpha(x|\mathcal{D}): \mathcal{X} \mapsto \mathbb{R}$ that assigns a score to any $x \in \mathcal{X}$, usually constructed from the surrogate $\hat{f}_{\mathcal{D}}$.
\item An \textbf{AF optimiser} that suggests the next decision point $x_{t+1}$, which yields the maximum of the AF.
\end{itemize}
Figure~\ref{fig:bo_framework} illustrates how these components integrate with ESA. In this work, we implement GP-UCB \cite{srinivasGaussianProcessOptimization2012}, a classic version of BO based on Gaussian processes (GPs) \cite{rasmussenGaussianProcessesMachine2008} and the upper-confidence-bound (UCB) \cite{lattimoreBanditAlgorithms2020} acquisition function.

GPs are the most widely used surrogate models in BO due to their flexibility in regressing arbitrary unknown functions with small datasets. Conceptually, a GP is a collection of random variables (indexed by $\mathcal{X}$, in our case), where any $n$ variables follow an $n$-variate Gaussian distribution. To specify a GP, we need a mean function $m: \mathcal{X} \mapsto \mathbb{R}$ and a covariance kernel $\kappa: \mathcal{X}^2 \mapsto \mathbb{R}$, denoted $\mathcal{GP}(m, \kappa)$:
\begin{equation}
\mu(x) = \mathbb{E} f(x),\quad \kappa(x, x') = \operatorname{Cov}(f(x), f(x')).
\end{equation}

\textbf{GP Prior and Initial Data.}  
The prior $P(f)$ is defined by selecting a prior mean $m(x)$ and prior kernel $k(x, x')$. A common choice is $m(x) = 0$. The prior kernel $k$ is chosen based on the properties of the decision space $\mathcal{X}$ and the available knowledge about the objective function $f$. For example, for a smooth $f$ on a closed set in Euclidean space $\mathbb{R}^d$, a plausible choice is the radial basis function (RBF) kernel:  
\begin{equation}
k(x,x') = c \cdot \exp \left\{-\|x-x'\|^{2} /{2 \ell^{2}}\right\},
\end{equation}
where $c$ and $\ell$ are hyperparameters. To acquire the initial dataset $\mathcal{D}_0$, we uniformly draw $m$ random parameters from $\mathcal{X}$ and evaluate their objectives.

\textbf{GP Posterior Surrogate.}  
Assume that each observation $y_t = f(x_t) + \epsilon_t$ is perturbed by $\epsilon_{t} \sim N\left(0, \sigma^{2}\right)$ i.i.d. Gaussian noise. Given that $P(f) \sim GP(m(x), k(x,x'))$ is the prior distribution of $f$, it can be derived that the posterior distribution of $f$ given $\mathcal{D}_t = \{(x_i,y_i)\}_{i=1}^t$ is also a GP, with mean function $\mu_t$ and kernel $k_t$. For simplicity, we define $\mathbf{y}_t := (y_1, \ldots, y_t)^\top$, $\mathbf{k}_t(x) := \left(k(x, x_1), \ldots, k(x,x_t) \right)^\top$, $\mathbf{m}_t := (m(x_1), \ldots, m(x_t))^\top$, and $\mathbf{K}_t := \left( k(x_i,x_j)\right)_{i,j}^{t \times t}$. The posterior distribution is given by \cite{garnettBayesianOptimization2023}:  
\begin{align}
\mu_t(x) &= \mathbf{k}_t(x)^{\top}\left(\mathbf{K}_t+\sigma^{2}\mathbf{I}\right)^{-1} (\mathbf{y}_t - \mathbf{m}_t),\\
\kappa_t(x,x') &= k(x,x')-\mathbf{k}_t(x)^{\top}\left(\mathbf{K}_t+\sigma^{2}\mathbf{I}\right)^{-1}\mathbf{k}_t(x').
\end{align}
Consequently, we have the surrogate $\hat{f}_t \sim \mathcal{GP}(\mu_t,\kappa_t)$. Note that in our current implementation, we only have one performance criterion, which is exactly the result of interest:
\[
\ell(y) = -y \in \mathbb{R},\qquad
f_{\pi}(x) = -\mathbb{E}_{y \sim \nu, \gamma \sim \pi(\cdot|x)}[\ell(y)]  = y.
\]
Therefore, this GP surrogate for $f_{\pi}$ also serves as the agent's model $\hat{h}$ of the experimental response.

\textbf{Acquisition Functions.}
The GP-UCB algorithm defines the AF as the upper confidence bound:  
\begin{equation}
\alpha(x|\mathcal{D}_t) = \mu_t(x) + \beta \sigma_t(x),
\end{equation}
where $\sigma_t(x) = \sqrt{\kappa_t(x, x)}$ is the posterior standard deviation, and $\beta$ is a hyperparameter. By setting a positive $\beta$, the UCB encourages exploration where the posterior surrogate is uncertain, while trading off the exploitation of high expected values.  

\textbf{AF Optimiser.}  
To balance computational efficiency and precision, we adopt a hierarchical grid search on the UCB acquisition function. It starts with a coarse grid of low resolution on $\mathcal{X}$, and recursively searches the regions near the high-AF points using a denser grid. The maximum of the AF is returned once a designated precision is reached. We also note that the decision space $\mathcal{X}$ must be discretised based on the manipulatory accuracy of hardware. Therefore, the final output of the AF optimiser is mapped to the closest point in the discretised $\mathcal{X}$.

\section{Pseudo Code of the BO Framework}
\label{app:bo_pseudo_code}
Algorithm~\ref{alg:bo} provides the pseudo code of the BO module used in this work. It summarises the iterative procedure described in Section~\ref{sec:bo}, including the initialisation of the search space, surrogate model fitting, acquisition function optimisation, and dataset update. The notation and variable names are consistent with those used in the main text for clarity.
\begin{algorithm}[H]
\caption{Bayesian Optimisation Framework}
\label{alg:bo}
\begin{algorithmic}[1]
\REQUIRE Search space $\mathcal{X}$, prior $p(f)$, observation function $\operatorname{obs}$ of the black-box objective $f$, initial points $\{x_i\}_{i=1}^m$, evaluation budget $T > m$, loss criteria $\ell$.  
\STATE Initialise $\mathcal{D}_0 = \{(x_i, y_i)\}_{i=1}^m$ by evaluating $y_i \sim \operatorname{obs}(x_i)$ 
\FOR{$t = m+1$ to $T$}  
\STATE Fit surrogate model to $\mathcal{D}_{t-1}$ to obtain posterior $p(f|\mathcal{D}_{t-1})$  
\STATE Optimise acquisition function: $x_t = \arg\max_{x \in \mathcal{X}} \alpha(x \,|\, \mathcal{D}_{t-1})$  
\STATE Observe $y_t = \operatorname{obs}(x_t)$
\STATE Augment dataset: $\mathcal{D}_t \leftarrow \mathcal{D}_{t-1} \cup \{(x_t, y_t)\}$  
\ENDFOR  
\STATE \textbf{Return} $x_{\text{best}} = \arg\max_{x_i \in \mathcal{D}_T} -\ell(y_i)$  
\end{algorithmic}
\end{algorithm}

\section{VPD Demonstration Dataset Collection}
\label{app:vdp-demonstration-dataset-collection}

We collected a demonstration dataset following a teach-by-demonstration paradigm. The resulting dataset, constructed to support a moderate level of generalisation, was employed to enhance the performance of the Visuomotor Policy Dictionary. In the first experimental scenario, the embodiment collected 170 trajectories per arm. Specifically, the left arm executed 20 demonstrations of grasping an electrode rack at a fixed position, along with an additional 150 demonstrations for generalisation, during which the rack was randomly placed within the workspace. The right arm generated 170 demonstrations encompassing a broad spectrum of electrode-grasping strategies. The trajectories of the left arm exhibit strong generalisation capability, whereas those of the right arm display greater diversity in trajectory morphology. In the second scenario, both arms recorded 70 trajectories each. Altogether, the two experimental settings produced a total of 480 demonstration trajectories.

We established a regular $10\text{ cm}\times10\text{ cm}$ grid over the plane reachable by the left arm, with each grid cell having a side length of 2~cm. Each $2\text{ cm}\times2\text{ cm}$ cell served as a subregion where the electrode rack was placed in 6 distinct poses, defined by both position and orientation, for demonstration collection.

Visual observations were captured using a monocular Logitech camera operating at 30~Hz, with each trajectory recording lasting approximately 30~s. Variability in visual scenes was introduced by randomising illumination conditions, the colour of the liquid in a beaker (transparent, light purple, or dark purple), and the electrode colour (pale yellow or light purple). In certain recordings, we deliberately introduced red and black electrode-wire motion as visual distractors to better emulate the visual noise commonly encountered in real laboratory environments. During data preprocessing, we further applied augmentations to the multi-view image set: approximately $1\%$ of the images were randomly replaced with pure black frames to encourage the model to rely on redundant visual cues from other views, thereby enhancing its robustness in generating coherent action chunks under partial visual occlusion.

\section{VPD Training and Inference Details}
\label{app:vdp-training-details}

\begin{figure}[H]
\centering
\includegraphics[width=\linewidth]{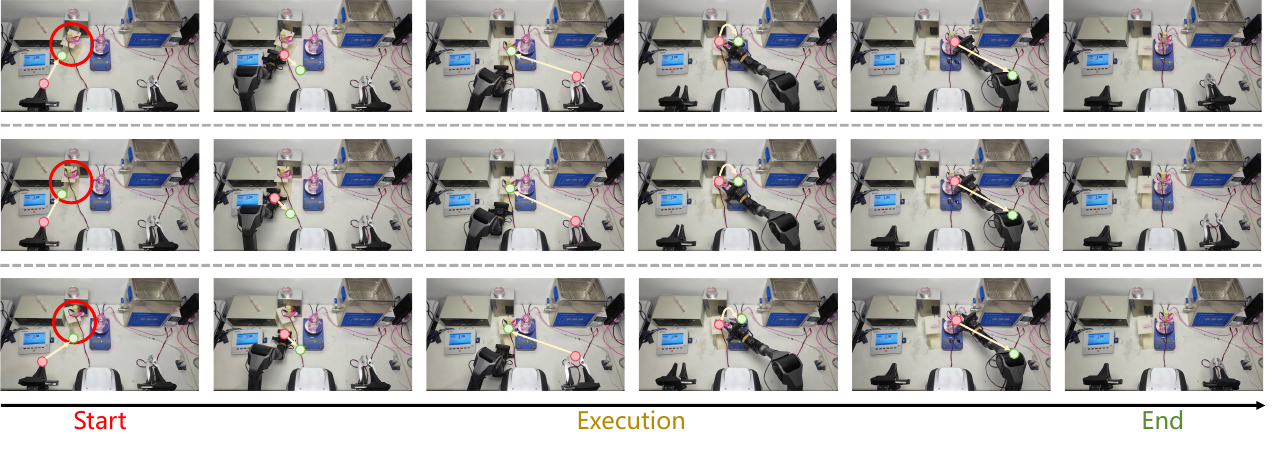}
\caption{Performances of the robot in the generalisation case study of Section~\ref{sec:vla}. The three rows correspond to three different initial placements of the electrode holder, circled with red boxes in the first column. Across all three trials, the robot's left arm reliably approaches and grasps the edge of the electrode rack with a reasonable grasping posture, while the right arm inserts the electrode into the beaker, demonstrating spatial generalisation of the learned visuomotor policies.
}
\label{fig:case1_gene_show}
\end{figure}

Training was conducted on an NVIDIA A100 GPU with 80 GB of memory. All models were trained from scratch with full parameter optimisation, while the pretrained visual backbone underwent full fine-tuning under the same learning rate. A batch size of 64 and an initial learning rate of $4\times10^{-5}$ were employed, with a cosine-annealing schedule applied to promote stable convergence. The training process lasted for approximately 200k steps, during which the validation loss converged to around 0.02 in the later stages, suggesting near-complete optimisation. The final checkpoint obtained at convergence was adopted for VPD evaluation.

Unless otherwise specified, all remaining hyperparameters followed the default configurations of \citet{zhao2023learningfinegrainedbimanualmanipulation}. To balance computational efficiency and model capacity, the number of layers in the modified Transformer encoders and decoders is set to 2. Empirically, this configuration demonstrated robust stability and generalisation, and it achieved consistency and accuracy in generating bimanual action sequences, particularly under multi-view input and multi-policy imitation conditions.

For the trained visuomotor policies, semantic labels are manually added to describe the tasks executable by each policy, thereby forming the VPD. These semantic labels are expressed in natural language and can be annotated using short phrases or complete sentences. During inference, retrieval can be performed using algorithms commonly employed in natural language processing, such as keyword matching or semantic similarity matching.

For the second long-horizon manipulation case mentioned above, frequent temporal synchronisation between the embodiment and the ESA is required to ensure that the embodiment can successfully perform time-sensitive chemical experimental operations. Consequently, the embodiment must promptly send feedback signals to the ESA upon the completion of each action. However, such feedback signals are often too rigid for conventional imitation learning methods, which typically rely on fixed time durations or predetermined task step counts.

To address this limitation, we propose a similarity-based mechanism for triggering the completion feedback signal. Specifically, we pre-obtain the embodiment's joint-angle vector $\hat{o}$ at the moment of task completion, and during inference, we continuously monitor the embodiment's current joint-angle vector $o_{t}$. The joint-angle similarity $s_{t}$ is then computed between the two: $$s_t = \frac{o_t \cdot \hat{o}}{\|o_t\|_2 \, \|\hat{o}\|_2}$$ When the similarity $s_{t}$ reaches a predefined threshold $s$, the task is considered complete, and the sub-policy within the VPD terminates inference. By continuously evaluating the joint-angle similarity $s_{t}$, the ESA can monitor the embodiment's task completion status and subsequently perform higher-level decision-making or reasoning processes.

\begin{figure*}[!th]
    \centering
    \subfigure[Module 1 and its corresponding chemical apparatus: funnel.]{
        \includegraphics[width=0.45\textwidth]{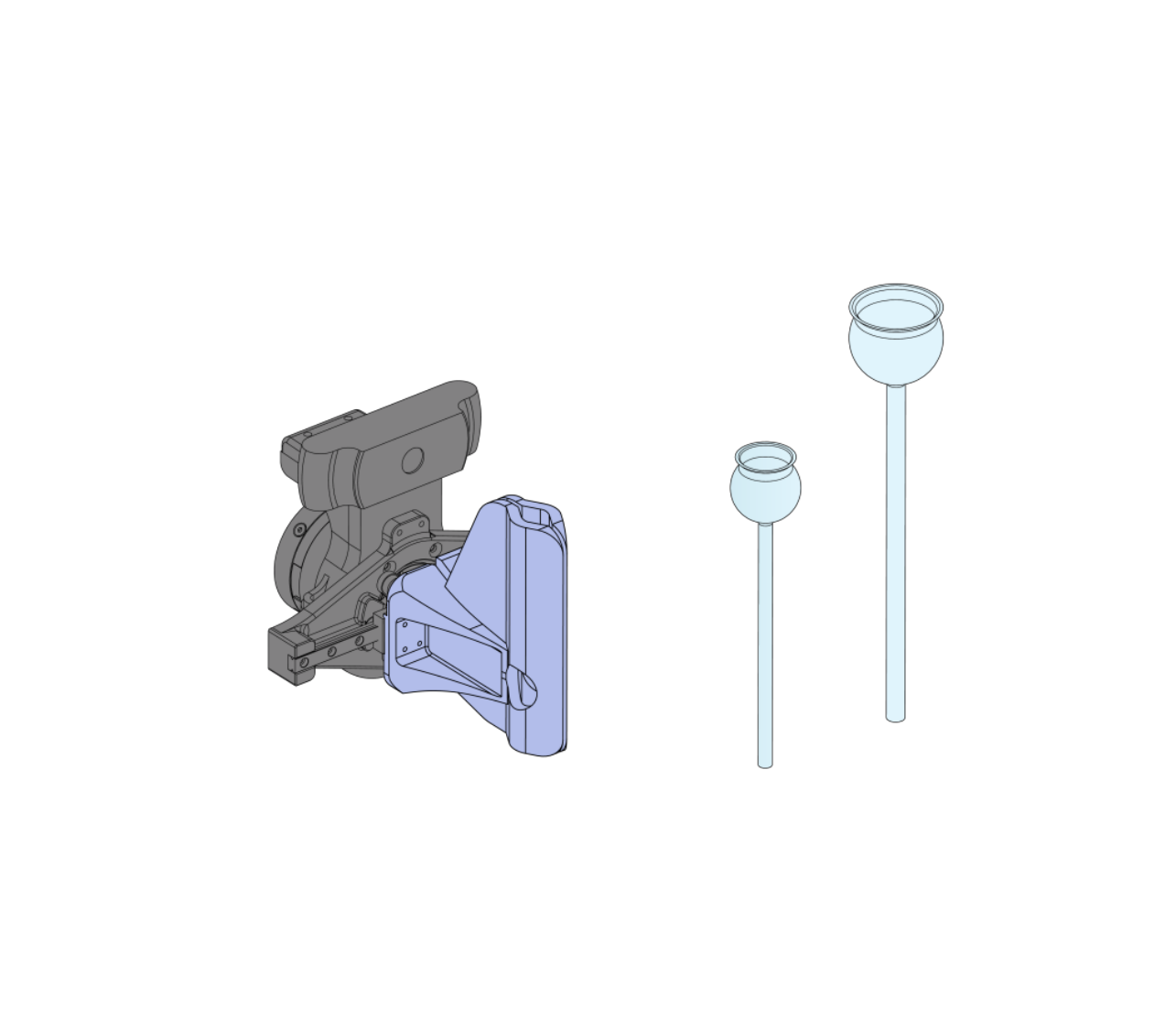}
        \label{fig:rsr_gripper_adapter_module_1}
    }
    \hfill
    \subfigure[Module 2 and its corresponding chemical apparatus: graduated cylinder, gas washing bottle with glass stopcock and its short type with volumetric flask with small size and large size.]{
        \includegraphics[width=0.45\textwidth]{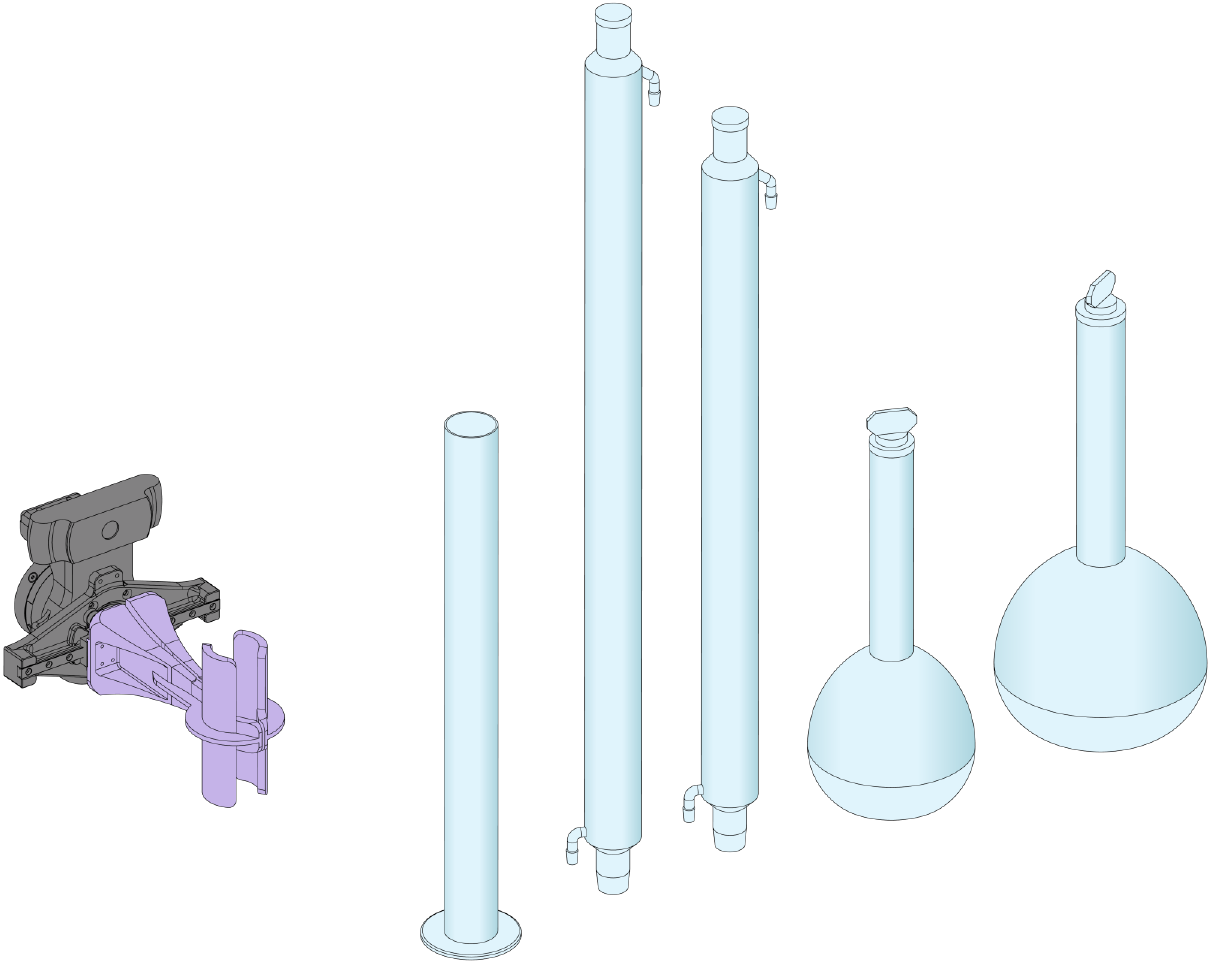}
        \label{fig:rsr_gripper_adapter_module_2}
    }

    \subfigure[Module 3 and its corresponding chemical apparatus: beaker with pour spout, watch glass and filter flask.]{
        \includegraphics[width=0.45\textwidth]{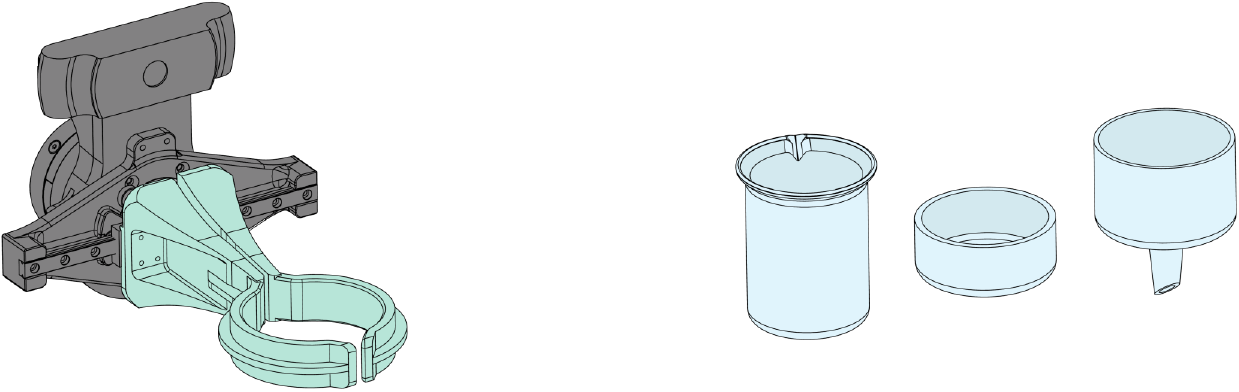}
        \label{fig:rsr_gripper_adapter_module_3}
    }
    \hfill
    \subfigure[Module 4 and its corresponding chemical apparatus: funnel, beaker, and reagent bottle.]{
        \includegraphics[width=0.45\textwidth]{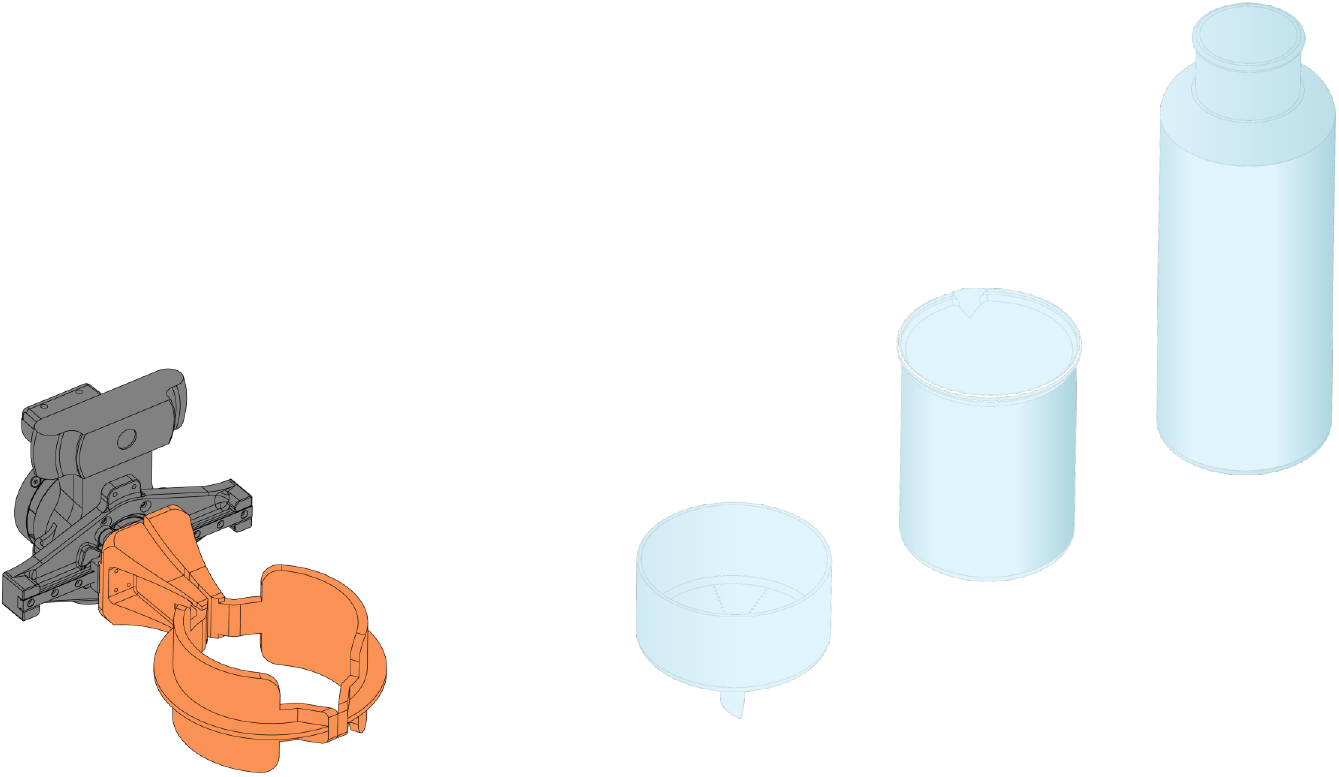}
        \label{fig:rsr_gripper_adapter_module_4}
    }

    \caption{3D prototype diagrams of the four modules and their corresponding compatible chemical apparatus.}
    \label{fig:rsr_gripper_adapter_all}
\end{figure*}
\section{Gripper Adapter Suite Details}
\label{app:gripper-adapter-suite-details}

We designed a gripper adapter suite for the embodiment, which enables the manipulation of various chemical laboratory instruments through simple configuration without the need to disassemble the original end-effector. This section presents the different adapter modules included in the gripper adapter suite and the corresponding instruments that can be grasped. Only four grippers are sufficient to handle a wide range of common laboratory instruments, eliminating the need for extensive hardware reconfiguration or software compatibility adjustments. 
Figure~\ref{fig:rsr_gripper_adapter_all} illustrates the gripper adapter suite and the corresponding laboratory instruments that each module is designed to manipulate.

\begin{table*}[t]
  \centering
  \footnotesize
  \caption{Detail of the Embodied Devices and Exposed Actions.}
  \label{tab:devices-actions}
  \setcellgapes{0pt}
  \makegapedcells
  \begin{tabular}{
    >{\centering\arraybackslash}p{3cm}
    >{\centering\arraybackslash}p{4cm}
    >{\centering\arraybackslash}p{1.5cm}
    >{\centering\arraybackslash}p{5cm}
  }
    \toprule
    \textbf{Device} & \textbf{Model} & \textbf{Interface} & \textbf{Actions} \\
    \midrule

    Magnetic Stirrer &
    IKA RCT Basic &
    API &
    \makecell[c]{\texttt{set\_temperature}, \\
                 \texttt{start\_heating\_for}, \\ \texttt{start\_stirring\_for}, \\
                 \texttt{set\_speed}, \texttt{reset}, \texttt{status}} \\

    \midrule

    Air Pump &
    Arduino Uno R3 &
    API &
    \makecell[c]{\texttt{set\_mode}, \texttt{start}, \texttt{stop}, \\
                 \texttt{start\_inflating\_for}, \\
                 \texttt{set\_speed}, \texttt{status}} \\

    \midrule

    Programmable Power Supply &
    QUAIQU SPPS3010D-232 &
    API &
    \makecell[c]{\texttt{power\_on}, \texttt{power\_off}, \\
                 \texttt{set\_voltage}, \texttt{set\_current}, \\
                 \texttt{get\_voltage}, \texttt{get\_current}, \\
                 \texttt{start\_output\_for}, \texttt{status}} \\

    \midrule

    Syringe Pump &
    ZHIGAO DP-03-60 &
    API &
    \makecell[c]{\texttt{initialise}, \texttt{switch\_channel}, \\
                 \texttt{aspirate\_ml}, \texttt{dispense\_ml}, \\
                 \texttt{transfer\_ml}, \texttt{status}} \\

    \midrule

    Spectrophotometer &
    INESA 721G &
    API &
    \makecell[c]{\texttt{measure}, \texttt{get\_transmittance}, \\
                 \texttt{get\_absorbance}, \texttt{status}} \\

    \midrule

    pH Meter &
    LEICI PHSJ-3F &
    GUI &
    \texttt{read\_ph}, \texttt{status} \\

    \midrule

    Electrochemical Workstation &
    Chenhua CHI660F &
    GUI &
    \makecell[c]{\texttt{start\_measurement}, \\
                 \texttt{wait\_for\_completion}, \\
                 \texttt{save\_results}, \texttt{status}} \\

    \midrule

    pH Probe &
    LEICI E-301-QC &
    VLA &
    --- \\

    \midrule

    Electrode &
    Gaossunion &
    VLA &
    --- \\

    \bottomrule
  \end{tabular}
\end{table*}
\section{Experimental Devices Details}
\label{app:experimental-devices-details}
\begin{figure}[tb]
\centering
\includegraphics[width=\linewidth]{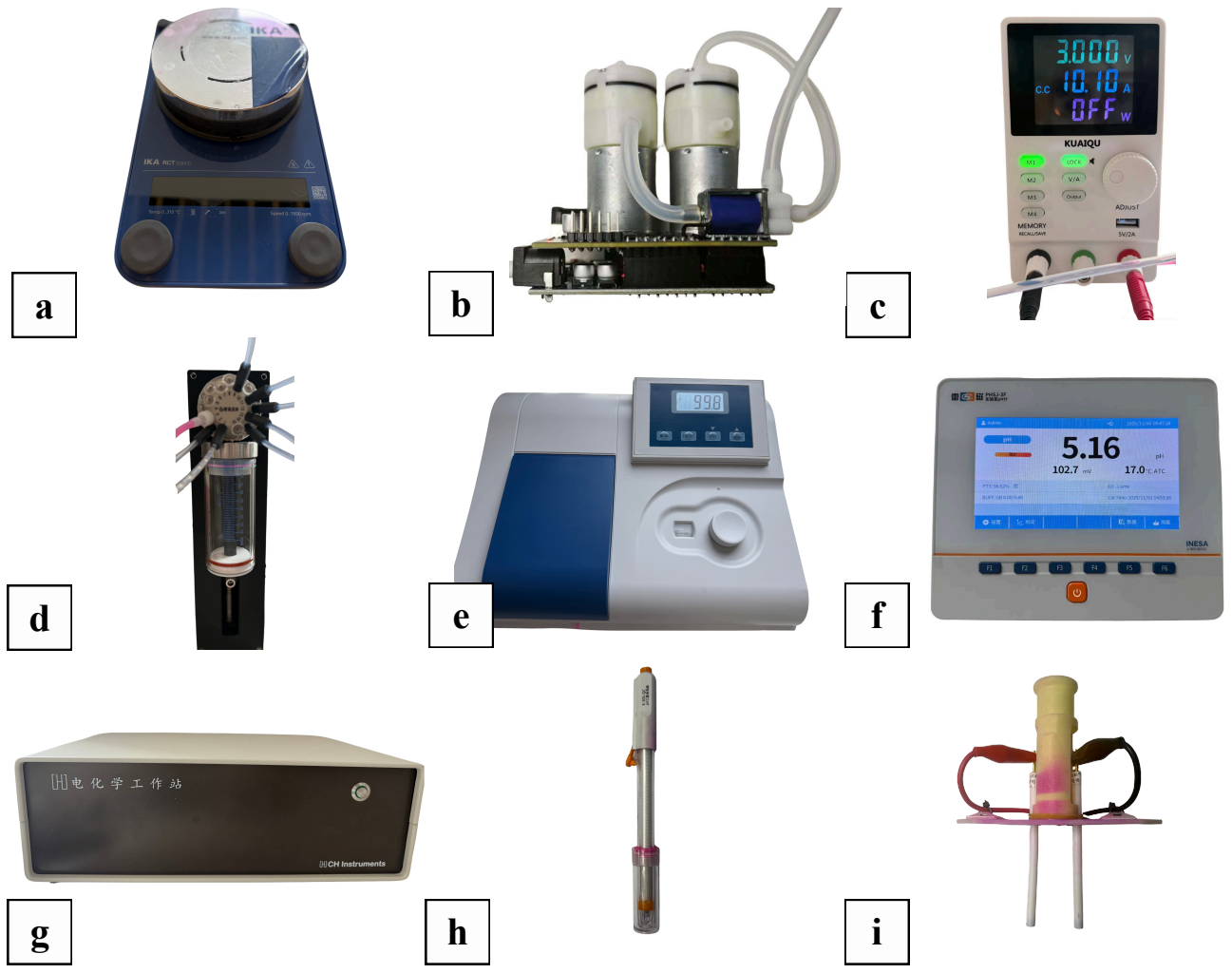}
\caption{%
Experimental devices used in our laboratory:
(a) IKA RCT Basic Magnetic Stirrer,
(b) Arduino Uno R3 Air Pump,
(c) QUAIQU SPPS3010D-232 Programmable Power Supply,
(d) ZHIGAO DP-03-60 Syringe Pump,
(e) INESA 721G Spectrophotometer,
(f) LEICI PHSJ-3F pH Meter,
(g) Chenhua CHI660F Electrochemical Workstation,
(h) Probe of the LEICI PHSJ-3F pH Meter,
(i) Custom Electrode Assembly.
}
\label{fig:exp_devices}
\end{figure}

This appendix provides detailed specifications and control interfaces of the experimental devices used in our automated laboratory platform (Fig.~\ref{fig:exp_devices}). Each instrument is connected through either an API-based or GUI-based interface, forming part of the EUI layer described in Section~\ref{sec:EUI}. Table~\ref{tab:devices-actions} summarises all devices and their exposed programmable actions.

\textbf{Magnetic Stirrer.}
The IKA RCT Basic Magnetic Stirrer is a device featuring an aluminium heating plate capable of reaching temperatures up to \SI{310}{\celsius}. For precise temperature regulation within the reaction medium, an external PT 1000.60 temperature sensor can be connected. The device is suitable for stirring volumes up to \SI{20}{\litre} (H\textsubscript{2}O). The top surface is made of toughened, shatterproof glass for chemical resistance and ease of cleaning. Control and data logging are facilitated through an integrated RS232 serial interface.

\textbf{Air Pump.}
A custom Arduino Uno R3 Air Pump aeration apparatus was developed based on an Arduino Uno R3 microcontroller. The system controls two air pumps and a three-way solenoid valve, enabling both aeration (gas pumping) and degassing (gas suction) modes. The rate of gas flow is precisely managed via serial commands using pulse-width modulation (PWM) to adjust the pump speed.

\textbf{Programmable Power Supply.}
The QUAIQU SPPS3010D-232 Programmable Power Supply is a DC power supply used in the experimental setup. It delivers a continuously adjustable output voltage (\SIrange{0}{32}{\volt}) and current with high stability and low ripple and noise. The unit includes integrated circuit protection features and is capable of sustained full-load output. It is controlled via its serial interface to set parameters and toggle the output.

\textbf{Syringe Pump.}
Fluidic operations were performed using a ZHIGAO DP-03-60 Syringe Pump. This high-precision device enables automated fluid transfer, dilution, and dispensing. Key features include an ultra-quiet motor, precise position feedback via an encoder, and optical limit detection. Communication with the device for aspirating, dispensing, and switching channels is handled via the Modbus protocol.

\textbf{Spectrophotometer.}
Qualitative and quantitative analyses were conducted using an INESA 721G Spectrophotometer. The instrument employs a holographic blazed grating monochromator to ensure high wavelength accuracy and low stray light. A microprocessor control system provides high-precision transmittance-to-absorbance ($T \rightarrow A$) conversion. Measurement initiation and data retrieval are performed via a serial communication interface.

\textbf{pH Meter.}
The pH of aqueous solutions was measured using a LEICI PHSJ-3F pH meter. The device supports 1- to 3-point calibration with automatic recognition of five standard buffer solutions and features automatic temperature compensation (ATC). As this device lacks a direct computer interface, readings were recorded manually through its graphical user interface.

\textbf{Electrochemical Workstation.}
The Chenhua CHI660F Electrochemical Workstation is a general-purpose electrochemical analysis system, which integrates a potentiostat with a fast digital signal generator for techniques such as AC impedance. It features a dual-channel high-speed data acquisition system and $iR$ compensation. Experiments were configured, initiated, and saved using the manufacturer-provided graphical user interface software.

\textbf{pH Probe.}
The LEICI E-301-QC pH triple-combined electrode is designed for the simultaneous measurement of solution pH and temperature. It is fully compatible with the LEICI PHSJ-3F pH meter, as well as ion meters, multi-parameter analysers, and titrators. The electrode features a durable plastic body that resists breakage, making it suitable for routine laboratory measurements.

\textbf{Electrode.}
We used a platinum sheet electrode from Gaossunion as the anode and used a composite electrode fabricated from waste Fenton sludge as the cathode. In addition, we designed a 3D-printed structural component to integrate the two electrodes, allowing convenient handling and gripping by the robot's right arm.

\end{document}